\documentclass[letterpaper, 10 pt, conference]{ieeeconf}
\usepackage{times}
\usepackage{multicol}
\usepackage[bookmarks=true]{hyperref}
\usepackage{graphicx}
\usepackage{multirow}
\usepackage{booktabs}
\usepackage{relsize}
\usepackage{xspace}
\usepackage{subcaption}
\usepackage{bbold}
\usepackage{listings}
\usepackage{bm}
\usepackage{bbm}
\usepackage{mathtools}
\usepackage{float}
\usepackage{soul}
\usepackage{textcomp}
\usepackage{gensymb}
\usepackage{tabularx}
\usepackage{makecell}
\usepackage{multirow}
\usepackage{hyperref}
\usepackage{booktabs}
\usepackage{xcolor}
\usepackage{pifont}

\usepackage{amssymb}% http://ctan.org/pkg/amssymb
\usepackage{pifont}% http://ctan.org/pkg/pifont

\newcommand{\tabref}[1]{Tab.~\ref{#1}}
\newcommand{\figref}[1]{Fig.~\ref{#1}}

\usepackage[color=todocolor, colorinlistoftodos]{todonotes}

\newcommand{\textjava}[1]{{\lstset{basicstyle=\ttfamily}\lstinline@#1@}}
\newcommand{\textjavafn}[1]{{\lstset{basicstyle=\footnotesize\ttfamily}\lstinline@#1@}}
\usepackage{setspace}
\usepackage{ifthen}

\long\def\sfootnote[#1]#2{\begingroup%
\def\thefootnote{\fnsymbol{footnote}}\footnote[#1]{#2}\endgroup}
\newcommand{\eg}{e.g., }
\newcommand{\ie}{i.e., }
\newcommand{\etal}{\textit{et al.}}

\newcommand{\ignore}[1]{}

\newcolumntype{M}[1]{>{\centering\arraybackslash}m{#1}}
\IEEEoverridecommandlockouts    % This command is only needed if you want to use the \thanks command

\begin{document}
% https://www.overleaf.com/project/60358f23403bdf18f9347844
% paper title
\title{GoferBot: A Visual Guided Human-Robot Collaborative Assembly System}

\author{Zheyu Zhuang$^{ \dagger1}$, Yizhak Ben-Shabat$^{\dagger 1,2}$,  Jiahao Zhang$^{1}$, Stephen Gould$^{1}$, Robert Mahony$^{1}$% <-this % stops a space
\thanks{ $\dagger$ Equal contribution.}
\thanks{$^{1}$The College of Engineering and Computer Science, The Australian National University, Canberra, Australia.
        {\tt\small \{FirstName.LastName\}@anu.edu.au}}%
\thanks{$^{2}$The Faculty of Electrical and Computer Engineering, Technion I.I.T, Haifa, Isreal 
        {\tt\small sitzikbs@technion.ac.il}}%
%\thanks{*This work was supported by The Australian Centre for Robotic Vision and The European Union MSCA-IF-GF fellowship ``3DinAction DLV-893465".}%
}   
\maketitle

\begin{abstract}

The current transformation towards smart manufacturing has led to a growing demand for human-robot collaboration (HRC) in the manufacturing process.
Perceiving and understanding the human co-worker's behaviour introduces challenges for collaborative robots to efficiently and effectively perform tasks in unstructured and dynamic environments.
Integrating recent data-driven machine vision capabilities into HRC systems is a logical next step in addressing these challenges. However, in these cases, off-the-shelf components struggle due to generalisation limitations. 
% These components are often evaluated on independent benchmarks. 
Real-world evaluation is required in order to fully appreciate the maturity and robustness of these approaches.
Furthermore, understanding the pure-vision aspects is a crucial first step before combining multiple modalities in order to understand the limitations.
In this paper, we propose GoferBot, a novel vision-based semantic HRC system for a real-world assembly task.
It is composed of a visual servoing module that reaches and grasps assembly parts in an unstructured multi-instance and dynamic environment, an action recognition module that performs human action prediction for implicit communication, and a visual handover module that uses the perceptual understanding of human behaviour to produce an intuitive and efficient collaborative assembly experience.
GoferBot is a novel assembly system that seamlessly integrates all sub-modules by utilising implicit semantic information purely from visual perception.
\end{abstract}

\IEEEpeerreviewmaketitle

\section{Introduction}
Automation of assembly and fabrication tasks has been common since the 1970s~\cite{singh2013robot_evolution}.
Limited by current robots' capability, human workers remain necessary for accomplishing complex cognition and manipulation tasks.
% In conventional factory settings fenced work cells are involved to enforce safety physical separation~\cite{shi2012levels_of_hrc}.
In the context of industry 4.0, human workers and collaboration robots are not only sharing the same physical workspace but also complementing each other's skill sets~\cite{billard2019trends,lasi2014industry}.
\begin{figure}
    \centering
    \includegraphics[width=0.42\textwidth]{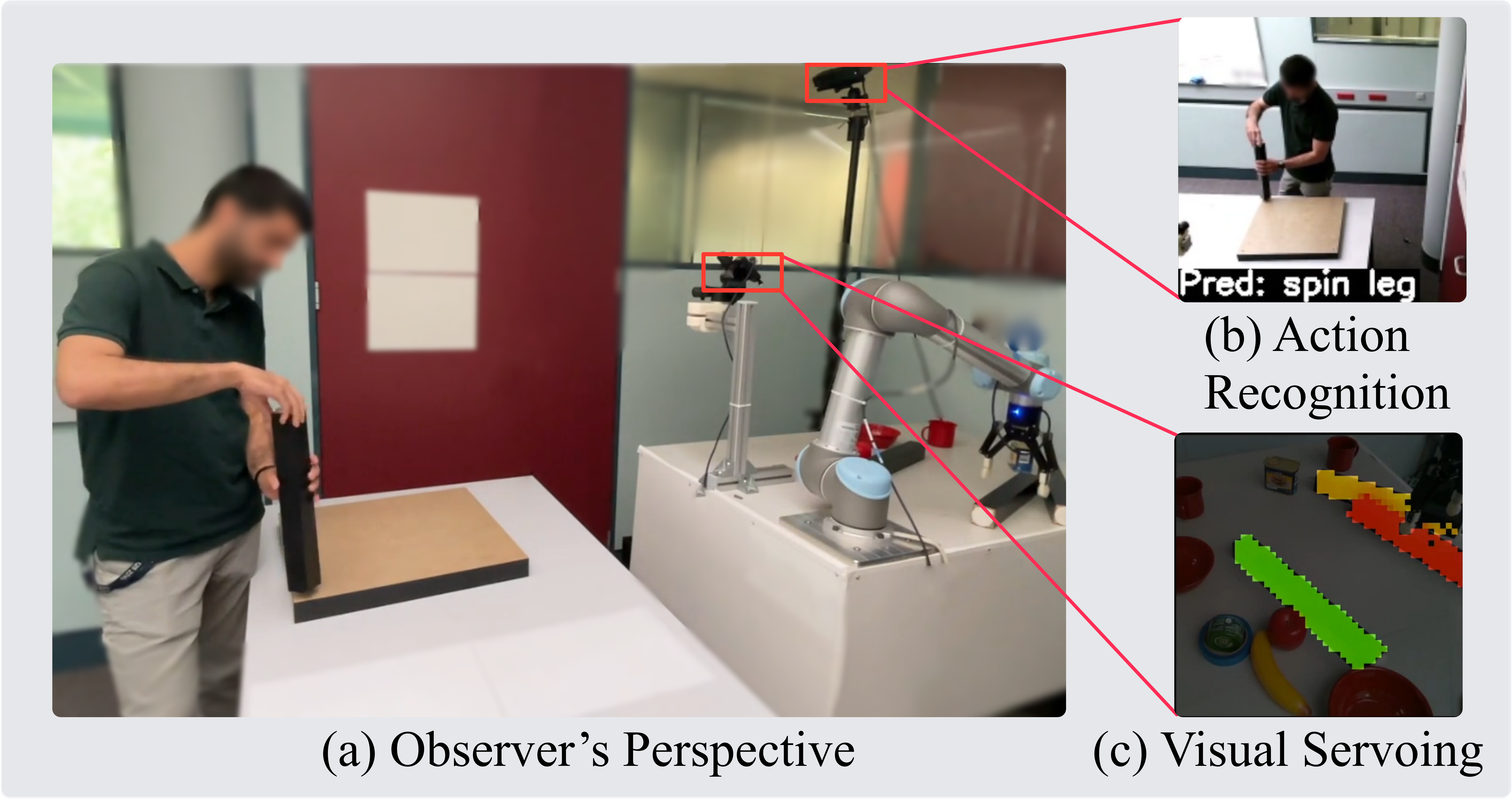}
    \caption{An illustration of the integrated assembly system. (a) An observer's perspective.
    including the human assembler, the assembly workbench on the left and the robot at the part storage on the right. 
    (b) The action recognition module view and the corresponding prediction.
    (c) The visual servoing module view with segmentation mask overlay, colour-coded according to regressed Lyapunov value.}
    \label{fig:overview}
    \vspace{-3mm}
\end{figure}
Towards the goal of achieving flexible, efficient and effective human-robot collaboration in an assembly task, a robot must: (1) deal with unstructured and dynamic environments, (2) perceive and understand human actions, (3) respond to human actions and task progress.
% \begin{enumerate}
% 	\item Deal with unstructured and dynamic environments,
% 	\item Perceive and understand the human actions,
% 	\item Respond based on human action and task progress.
% \end{enumerate}
The first two challenges are strongly tied to the research of human action recognition/prediction and visual-guided robotic manipulation.%, and scene understanding.
These areas are advancing as a result of machine learning developments.
Recently, data-driven approaches have been proposed in human-robot collaboration settings such as identifying/predicting industry-related actions~\cite{ZHANG20209} and gesture recognition~\cite{xia2019vision}, \textit{etc.} 

As is frequently the case in the machine-learning community, these real-world oriented algorithms are commonly evaluated against datasets.
Although task-specific benchmarks are valuable metrics, it is critical for a system to be evaluated as a whole in real-world scenarios \cite{ZHANG20209}. 
% To bring the proposed industry-oriented learning module one step closer to real-world evaluation, work such as \cite{ZHANG20209, michalos2018industral_hrc} proposed fully integrated systems.
In addition, a data-driven HRC system must explicitly take human experience such as fluency and ease-of-use into account rather than relying solely on robot-centric metrics for evaluation. 
These goals will lead to more substantial evaluation and verification of integrated systems for integrated manipulation in human-machine collaboration \cite{michalos2018industral_hrc}.
% Beyond being evaluated against robot-centric metrics, a data-driven HRC system must take human experience into account, such as fluency and ease-of-use.

Due to the rapid growth of vision-based learning, it is worthwhile to investigate the raw capability of a vision-only, data-driven HRC assembly system as well as the human coworker's experience. 
This not only provides direction for improving or adapting state-of-the-art learning sub-modules for real-world HRC systems from robot-centric and human-centric perspectives but also serves as a reference for later vision-based, data-driven, systems prior to incorporating additional sensing modalities and redundancies.

This paper presents a novel human-robot collaborative assembly system that demonstrates all of the aforementioned capabilities and uses vision as its only exteroceptive sensing modality. 
The robot proactively reaches and grasps an assembly part from multi-instance and dynamic environments, and presents the part until a human grasp action is visually perceived.
A key attribute of the system is that the robot naturally interacts and adapts to human actions, providing an intuitive experience for the human, which resembles the synchronised coordination among human workers developed during instructional fabrication and assembly tasks.

%We use vision as the core sensing sensing modality, providing both visual action perception for the human, and visual servo control of the robot to grasp and pass parts to the human.
%The visual action perception system uses an external front-top view monocular camera to predict human assembly action as shown in Fig.1a.
%The visual control system robustly reaches and grasps a piece amid multiple identical instances and distractors via the second monocular camera positioned to observe the working-bench (Fig.1b).
%The robot-to-human hand over task is implemented as a timed coordinated action between robot and human, where the natural reaching action of the human is recognised by the action perception system and the robot coordinates the passing action and release of the object based on the visual confirmation of a established grasp.

The system is particularly of interest since its perception module does not rely on depth sensors, fiducial markers, or motion trackers, requires only two RGB cameras; one that observes the human and one that the robot uses for the visual servoing task (\figref{fig:overview}).
The simplicity of the setup and robustness of the resulting algorithm are important contributions of the work in demonstrating the potential of such data-driven vision-based human-robot collaborative systems for flexible assembly tasks.
These characteristics are ideally suited to provide shared autonomy for a wide range of tasks in unstructured and open environments.

% The main contribution of this work is the demonstration of a fully integrated data driven, vision-based human-robot collaborative system for assmebly. 
% First, develop a fully integrated vision-based human-robot collaborative system for assembly. Second,  we propose human-inspired implicit semantic communication to enable efficient human-robot collaboration.
% The contributions are supported by extensive human-centric and robot-centric evaluation of the proposed fully integrated end-to-end system. 

% In particular, the system comprises: 
% \begin{enumerate}
%     \item An action perception system that understands and predicts human action to enable natural human-robot interaction
%     \item A visual servo system that grasps target items from unstructured and cluttered workspaces.
%     \item A human-robot collaborative control that passes objects to a human in a natural and intuitive manner.
% \end{enumerate}

%the capability of vision sensors in through a purely vision based human-robot collaboration assembly task.
%The robot reach and grasp pieces from an unstructured tabletop, pass and hold the piece at a designated pose, waiting for the human-worker to reach and grasp. Once the worker has grasped the piece, the robot releases the piece and move on to collecting the next.
%only two monocular cameras are deployed. The interaction, handover, are all handled by monocular vision.

% We evaluate the performance of each module separately using acceptable accuracy metrics and the system as a whole using objective and subjective metrics. 
% \zheyu{expand this}

\begin{figure*}
    \centering
    \includegraphics[width=0.85\linewidth]{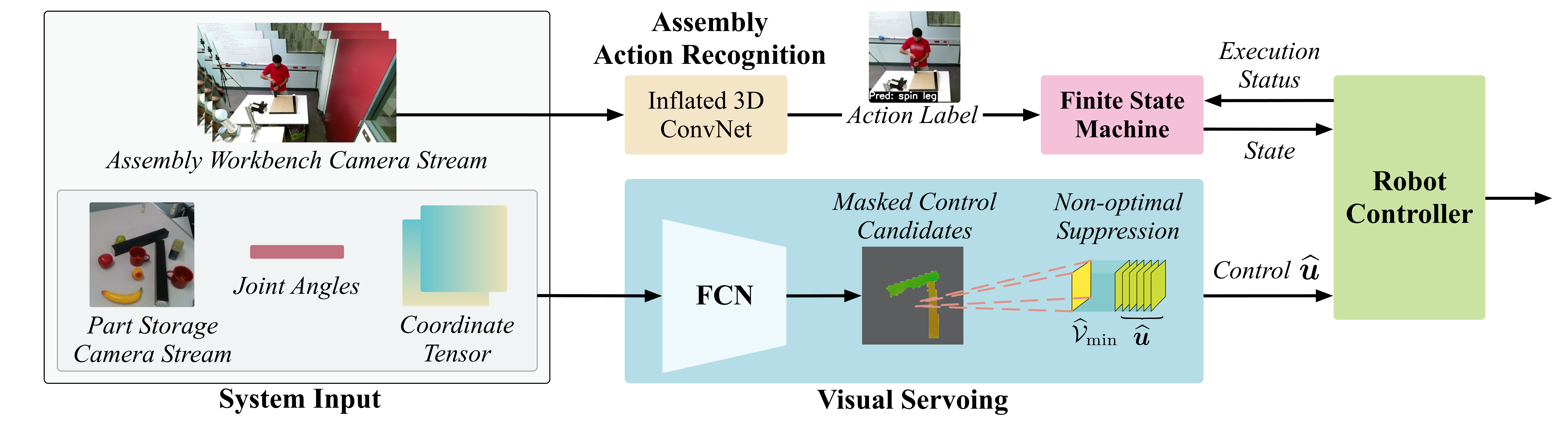}

    \caption{GoferBot full system illustration. The action recognition module gets as input an image sequence of the human and predicts the human action and then passes it to a finite state machine (FSM). The robotic visual servoing module gets the storage-space image sequence along with the current robot joint angle and normalized image pixel coordinates and passes the inferred control to the robot controller. The robot controller then gets the state from the FSM and translates both inputs to joint control. Once assembly finishes, the controller passes an execution status to the FSM.}
    \label{fig:approch_block_diagram}
    \vspace{-2mm}
\end{figure*}

\section{Related Work}
In contemporary manufacturing settings, a human-robot collaboration robot should be capable of autonomously performing low-level manipulation tasks such as grasping or non-prehensile manipulation, perceiving human intents via gaze, gesture, or action recognition/prediction, and interacting with a human coworker based on the perceived intent in real-time.
This section reviews related work on the three aforementioned aspects considered in this work, with an emphasis on modern data-driven, vision-based techniques.

\subsection{Video Perception}
Explicit communication such as gesture~\cite{nickel2007hri_gesture, gleeson2013gestures} and auditory commands~\cite{petit2012hri_language} essentially requires the human to adapt to the robot instead of forming intuitive collaborations with the robot co-worker.
In recent years, the computer vision community has devoted much attention to the task of action recognition. This has manifested in a plethora of large-scale action recognition datasets (\eg Kinetics~\cite{Qiu_2017_ICCV},  Epic Kitchens~\cite{damen2018scaling}). %ActivityNet~\cite{caba2015activitynet}), 
%as well as task specific and instructional datasets (\eg  Epic Kitchens~\cite{damen2018scaling},
%COIN~\cite{tang2019coin}
%, and YouCook~\cite{das2013thousand, zhou2018towards}.
Recently, a couple of datasets have focused on the task of furniture assembly like the IKEA-FA~\cite{toyer2017human} and IKEA ASM datasets~\cite{ben2021ikea}. While each dataset provides different information modalities, they all include a sequence of RGB frames along with a per-frame action annotation. 
    
Current state-of-the-art action recognition methods are based on deep convolutional neural network (CNN) architectures that process video data. The most prominent approaches extract spatio-temporal features using 3D convolutions. The Convolutional 3D (C3D)~\cite{tran2015learning} method was the first to apply 3D convolutions in this context, and was followed by pseudo-3D residual net (P3D ResNet)~\cite{Qiu_2017_ICCV}, which utilises residual connections, simulates 3D convolutions and essentially leverages pre-trained 2D CNNs. Similarly, the inflated 3D ConvNet (I3D)~\cite{carreira2017quo} uses an inflated inception module and provides an action class prediction for each frame.
 
All of the presented methods were used in a controlled off-line benchmark setting which does not take into account several critical real-life challenges that affect performance when integrated into a robotic system, such as input frame-rate, and the restriction of using only past frames as input. The above methods' design and evaluation focus on action recognition after the fact while in real-time robotics systems the focus is on preemptive action recognition. Furthermore, the above methods exist and are evaluated in isolation and are disconnected from any robotics system where there is higher importance for the system performance as a whole. 
 
    % more details on I3D ? here ? Not enough space!
In this paper, we extend the I3D architecture for real-time action recognition as a part of the GoferBot system. The network was pre-trained on Kinetics and IKEA-ASM dataset and fine-tuned using data we have collected. This method was chosen due to its ability to operate in real-time while maintaining good action prediction accuracy.

\subsection{Visual Servoing \& Grasping}
% Controlling the end-effector of the manipulator relative to a target from visual information, i.e. visual servoing, has been a topic of great concern in robotics since the 1980s~\cite{corke1993visual, hutchinson1996tutorial}. 
% There are two fundamental paradigms in the literature: position-based visual servoing (PBVS) and image-based visual servoing (IBVS). The former firstly estimates the position or pose of the target object(s) from the image and use an off-the-shelf position-based controller servoing the manipulator. The latter directly derives the control action from the image input.

In the deep learning era, performing robotic grasping based on object pose estimation is a common strategy.
Despite continuous advancement in object pose estimation algorithms~\cite{peng2019pvnet, zakharov2019dpod, he2020pvn3d, wang2019densefusion}, comparatively fewer learning-based pose estimation algorithms have been integrated into real-time, closed-loop robotic grasping systems due to domain gap and computational expensive inference.
% , the majority of the learning-based pose estimation algorithms are not suited for a real-time closed-loop system.
Tremblay \etal~\cite{tremblay2018deep} propose the first pose estimation algorithm that is integrated into a multi-instance and real-time grasping system ($\sim$10 fps).
% Later, Tremblay \etal~\cite{tremblay2020indirect} combine this pose estimation algorithm with an online camera-to-robot estimation network pipeline so that the external camera can be freely moved whilst the robot performs grasping actions.`
An alternative to object pose estimation is grasp synthesis~\cite{mousavian20196rgb_grasping, murali20206rgbd_grasping_clutter}, which directly identifies grasp configurations based on the geometric information of the target.
% Mousavian \etal~\cite{mousavian20196rgb_grasping} generate a set of 6 DoF grasp proposals based on segmented point clouds. Murali \etal~\cite{murali20206rgbd_grasping_clutter} extend this approach by learning to discard grasp proposals that collide with the environment.
In the context of dynamic environments, GG-CNN \cite{morrison2018ggcnn} uses depth images to predict top-down, object independent, grasp proposals in real-time.
% Rosenberger \etal~\cite{rosenberger2020realtime_handover} propose an object-independent real-time human-to-robot handover system by incorporating the GG-CNN into the system.
% with a human body part segmentation, an object detection and a hand/ﬁnger segmentation pipeline.
Grasp synthesis algorithms rely heavily on the quality of the geometric information.
However, the accuracy of a depth sensor is compromised when the object surface is reflective or absorptive.
% , e.g. the dark IKEA furniture piece.

Learning end-to-end visuomotor policy is an alternative to the pose-based approaches.
James \etal~\cite{james2017transferring}, Rusu \etal ~\cite{rusu2017sim} and Zhang \etal~\cite{zhang2019adversarial} demonstrate image-to-control networks servoing the manipulator to reach (and grasp) simple single geometric shapes using RGB image inputs.
Hämäläinen \etal~\cite{hamalainen2019affordance} propose a modular network architecture with separate perception, policy and trajectory blocks reaching towards a single instance amid distractors.
Reinforcement learning is also explored for visuomotor policy learning, such as for bin picking~\cite{levine2018learning, kalashnikov2018qt}, and learning from human demonstrations~\cite{smith2019avid}.
In manufacturing settings, it is common having more than one object instance, such as assembly parts, appearing within the same image frame. 
A Lyapunov function motivated multi-class, multi-instance reaching network (LyRN) is proposed in~\cite{zhuang2021multi}.
It reaches and grasps a variety of objects in an unstructured and dynamic environment with up to 160Hz closed-loop control.
In this work, we exploit and extend the highly reactive, multi-instance, visual servo control developed by the authors in prior work~\cite{zhuang2020lyrn, zhuang2021multi}. 

% This paper extends the highly reactive, multi-instance, visual servo control developed in prior work~\cite{zhuang2020lyrn, zhuang2021multi}. 

% \subsection{Human-Robot Collaboration Systems}
% Creating complete human-robot collaboration systems is sophisticated, yet rewarding.
% Human-robot collaboration systems are cross-disciplinary and  structurally complex.
% Instead of tackling sub-tasks in isolation, many researchers study different aspects of human-robot collaboration systems, such as task modelling, coordination strategy and handover under the context of integrated HRC systems, providing deeper and more accurate evaluation.
% Due to the complexity of constructing a complete HRC system, simplifications and relaxations are made at non-essential sub-modules.
% For example, \cite{Unhelkar.2018} focus on trajectory forecasting by suing off-the-shelf Kinect and OpenNI depth modality for human tracking and pose estimation. \cite{Roncone.2017} intervenes with the environment by using Aruco markers for perception.
% \cite{huang2015adaptive} studies the adaptive handover strategies in a complete HRC framework with a robot load dishes and passes them to the human co-worker. 
% However, this systems uses hard-coded robot grasping strategy and triggers release based on force-torque sensor, which is prone to external perturbation.

\subsection{Robot-to-Human Handover}

GoferBot aims at minimising the human idle time during an assembly task. Under the current technology capability, static handover, where the robot passes the part to a designated handover location and waits for human pickup is the most efficient.
Thus, one of the main focuses of the system is to determine the time of release.
One well-studied method is monitoring and modelling changes in a force/pressure sensor~\cite{huang2015adaptive, konstantinova2017autonomous_handover_tactile}.
However, due to the simplicity of the force sensor, it has limitations in filtering out induced forces from the environment~\cite{ajoudani2018hriprogress}. 
Grigore \etal~\cite{grigore2013joint_gaze}, Admoni \etal~\cite{admoni2014deliberate_gaze} use the human gaze as additional redundancy to robustify the handover decision.
% In this work, we directly estimating the time of release using an external monocular camera that observes human actions.
Apart from the time of release, there are other challenges of a robot-to-human handover task, for example, estimating handover location, motion planning and control.
We refer the readers to~\cite{ortenzi2020handover_review} for a comprehensive review on robotics handover.
In this work, we approach the robot-to-human handover sub-task as a vision problem by estimating the human reach and grasp action from the same RGB camera that observes the human coworker. 
% visual handover

\subsection{Full HRC Systems and Evaluation Metrics}
Having a full system with integrated grasping and handover/delivering capability is crucial for studying various aspects of HRC systems, such as trajectory estimation, task modelling, coordination strategy, human-centric studies, \textit{etc.}
% In a human-robot handover collaboration system, it is crucial for a robot to hand objects to a human reliably.
Due to the complexity of establishing sophisticated full systems,  simplifications and relaxations are often made by using off-the-shelf sub-modules to reduce the overall complexity in order to focus on improving specific aspects of the system.
For example, Unhelkar \etal~\cite{Unhelkar.2018, ZHANG20209} use Kinect and depth modality for human tracking and pose estimation, Roncone \etal~\cite{Roncone.2017} intervene with the environment by using Aruco markers for perception, and Huang \etal~\cite{huang2015adaptive} use a hard-coded robot grasping strategy and trigger object release based on force-torque sensor readings.
% These systems are designed to study specific aspects of the HRC system yet not geared towards real-world applications.
Extensions and redesigns for simplified sub-modules are required for deploying them in the real world.
% In this work we push the system one step closer to a real world scenario by extending, adapting, and fine-tuning a data-driven approach for vision-based perception and relaxing only the task modelling aspect.

\textbf{Human-centric evaluation metrics: } The evaluation of an efficient and fluent robotic handover has been studied extensively as a human-robot joint action. Prominent metrics measure objective and subjective parameters that vary according to the system and task. Hoffman~\cite{hoffman2019evaluating} has developed metrics to evaluate human-robot teamwork level of fluency in a turn-taking framework. Subjective metrics measure the level of trust, robot contribution and the human sense of the robots' commitment to the team while the objective metrics measure the percentage of human and robot concurrent activity (C-ACT), human idle time (H-IDLE), the robot's functional delay and the robot's idle time (R-IDLE).  These metrics were developed in a specific collaborative scenario and have built-in, task-specific assumptions that may not hold for other scenarios. Variations of Hoffman's metrics appeared in previous work, specifically relating to handover, including Huang \etal~\cite{huang2015adaptive} that studied handover of dishes and showed that while time performance improved, there was a negative effect on fluency, Cakmak \etal~\cite{cakmak2011using} that used spatial and temporal contrast, and  Han and Yanco~\cite{han2019effects} that explored handover proactive release behaviours. 

% More specific to the task of handover,  Huang \etal \cite{huang2015adaptive} used similar metrics for the task of human-robot handover in unloading a dish rack while using objective fluency metrics (C-ACT, H-IDLE, R-IDLE) and completion time. They showed that while time performance improvement can be achieved using proactive coordination, it effects negatively on the fluency. They also show an inverse result for reactive coordination which indicates a non-trivial relation between objective and subjective fluency. In a different study by Cakmak \etal \cite{cakmak2011using}, fluent robot to human handovers were developed using spatial and temporal contrast. They conducted a survey to subjectively evaluate reliability in addition to an experimental study for objective fluency evaluation using human and robot functional delay. Han and Yanco \cite{han2019effects} have explored proactive release behaviours during human-robot handovers by measuring objective completion time and a subjective questionnaire that evaluated aspects of fluency, trust, comfort, capability, and ease of use.

% In this paper, we adopt the evaluation metrics proposed by Hoffman \cite{hoffman2019evaluating} and adapt them to the full assembly task and to the specific handover module. We evaluate using the objective measures of R-IDLE, H-IDLE, C-ACT and cycle completion time as well as the subjective measures of fluency, trust, comfort, capability, and ease of use. 

\section{GoferBot}
\begin{figure*}
    \centering
    \includegraphics[width=0.92\linewidth]{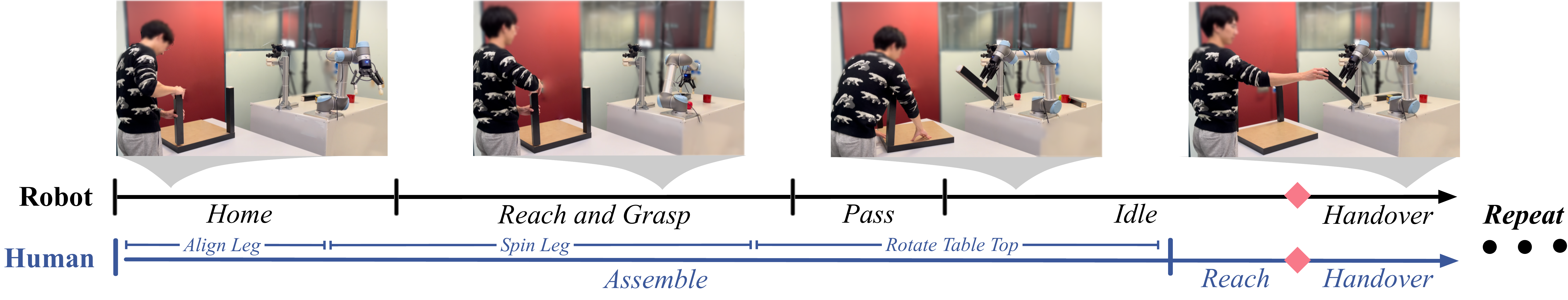}

    \caption{An illustration of a human-robot leg assembly cycle. The human performs an ``assemble" action which is composed of multiple atomic actions. During that time, the robot transitions between, home, reach and grasp, and pass states and then waits for the human to complete the assemble action in ``idle" state. The pink diamond symbol denotes the time of synchronisation between the robot and the human and the start of the collaborative handover.}
    \label{fig:assemble_steps}
    \vspace{-2mm}
\end{figure*}

GoferBot is a vision-based human-robot collaborative system for an assembly task. The assembly task explored in this paper refers to an assembly of an IKEA Lack table. The full assembly process is composed of four single-leg assembly cycles. For the human, each cycle is composed of receiving a leg from the robot, aligning the leg to the designated hole, spinning the leg in and then optionally rotating the tabletop (in preparation for the next cycle) or flipping the table once all four legs have been assembled. For the robot, each cycle is composed of homing, reaching and grasping a leg in the part storage area, passing to the delivery point at the assembly workstation and then waiting for the human to grasp the leg to initiate a handover. A single cycle is illustrated in \figref{fig:assemble_steps}.
% (See supplemental material for GoferBot's collaborative assembly video). 

To seamlessly achieve this task, GoferBot comprises three sub-modules: action recognition, visual servoing, and visual handover. These modules operate simultaneously, receiving video streams from two RGB cameras and extracting semantic information.
A block diagram illustrating the system is depicted in \figref{fig:approch_block_diagram}.
% Specifically, the action recognition module performs action prediction and passes the current action to a finite state machine. At that time, the visual servoing module grasps an assembly part and passes it to the delivery point where the visual handover module is initiated when a human grasp action has been identified. 
% We provide additional details on each module in the following subsections.
Note that task modelling is not a part of the proposed system. 

\section{Hardware Setup}
% \subsection{Hardware Setup}
GoferBot and its workspace are physically separated into two sectors: part storage and assembly workstation (\figref{fig:overview}).
The human needs the robot's assistance to transfer suited parts from the storage to the assembly workstation.
Video RGB streams from a Kinect V2 and an Intel RealSense D455 are used as the system's vision inputs.
The Kinect v2 is positioned from an angled top-down view, overseeing the assembly workstation in order to recognise human action; the RealSense is mounted on the table surface to the storage for performing vision servoing and grasping.
The depth information perceived by these sensors is discarded.
% While these sensors can also provide depth information, in this paper we only use their RGB streams. 

The robot is a six degrees of freedom UR5, mounted alongside the RealSense on the storage table. The end-effector of the robot is Robotiq 140mm parallel gripper with custom textured silicone fingertips.
% The vision system inference is done using an Nvidia GeForce GTX 1080 Ti GPU on an Intel i7-8700K CPU with 32GB RAM machine. 
% \zheyu{More details of the vision system here?}
% \IBS{ I added the machine info, is that what you mean?}
The system's computation is distributed over two computers with a single 1080Ti GPU in each. Each computer separately processes a single video  --- the Kinect v2 for action and RealSense for visual servoing.
Hardware messaging is handled by ROS.

\subsection{Assembly Action Recognition}
Human actions for furniture assembly are composed of multiple atomic actions. Some actions are instrumental for the assembly process while others are either optional or dependent on the assembler's choices. 
For example, assembling the IKEA Lack table consists of attaching four legs onto a tabletop, however, the order of leg assembly is arbitrary and occasionally the tabletop will be flipped or will require rotations for ease of assembly. 
Furthermore, having a robot in the loop eliminates some of the actions previously done by the human (\eg picking up a leg) while introducing new actions for the human-robot interaction (\eg human reaching and grasping parts). Therefore, the proposed action recognition module is composed of a SOTA neural network and designated data, collected particularly for this task. 

\begin{figure}
    \centering
    \begin{subfigure}{.49\linewidth}
        \centering
        \includegraphics[width=0.99\linewidth]{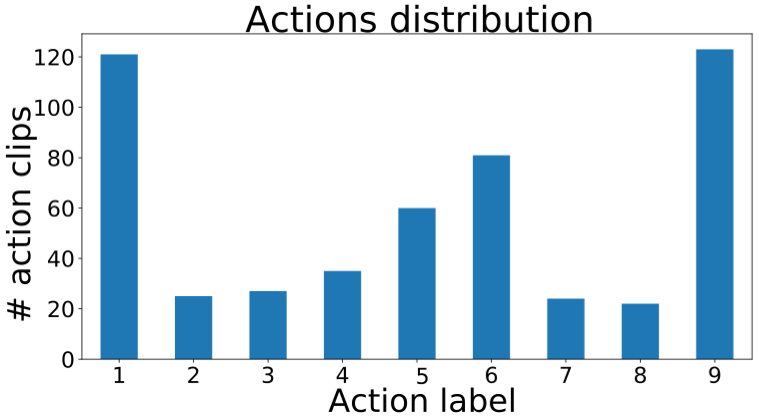}
    \end{subfigure}
    \begin{subfigure}{.49\linewidth}
        \centering
        \includegraphics[width=0.99\linewidth]{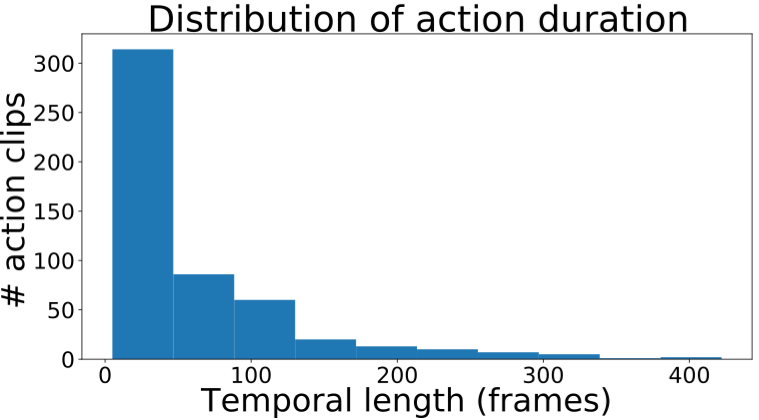}
    \end{subfigure}
    %\begin{subfigure}{.32\linewidth}
    %     \centering
    %     \includegraphics[width=0.99\linewidth]{figs/action_occ_dist_.png}
    % \end{subfigure}
    \caption{Human-robot collaborative action dataset statistics. Distribution of: action clips (left), and action duration (right)}%, and Action occurrence (right).}
    \label{fig:dataset_stats}
    \vspace{-2mm}
\end{figure}

% \noindent
\textbf{Data: } In this work, we extend the IKEA ASM dataset~\cite{ben2021ikea} to the human-robot collaborative task. The atomic actions used in this work are: 1. No assembly action,  2. reach (to grasp assembly part), 3. flip tabletop, 4. flip table, 5. spin leg, 6. align the leg, 7. rotate table top, and 8. grasp (assembly part).  To predict this new set of actions we collected six new assembly videos. In contrast to the IKEA ASM dataset, these videos are not of full assemblies but rather of multiple repetitions of the above atomic actions, where a human teleoperates the robot to simulate the hand-over of assembly parts. Each video portrays a different individual performing the atomic actions in their natural, unscripted way. The videos are then manually annotated with atomic actions and used in training as short 16-frame clips. Therefore at training time, an end-to-end assembly is never fully seen. The statistics of the collected data are given in \figref{fig:dataset_stats} and show the high imbalance of action duration (\eg spinning a leg takes much longer than grasping a leg), as a consequence some actions have more clips available during training. At training time we mitigate this imbalance using a weighted data sampler. When collecting the data, the action occurrences (number of action instances in each video) was approximately balanced within the range of 5--6 occurrences and an average of 5.8 action occurrences per video.

% \noindent
\textbf{Inference: } For action inference we use a single stream I3D architecture \cite{carreira2017quo}. 
It uses inflated inception modules and learns spatio-temporal filters. 
The network is initialised from pre-trained weights on IKEA ASM and its final layer is modified to match the eight atomic actions. 
To enable real-time processing, we choose the single stream model since it has much fewer parameters than the two-stream model and does not require computing optical flow.
The I3D architecture receives as input a sequence of $224 \times 224$ cropped frames from the Kinect V2 stream.
Since some actions' duration are very short, we use a sequence length of 16 frames for inference. When launching the action recognition, the first 15 frames are buffered and inference is enabled from the $16^{th}$ frame onward at a rate of ten frames per second. In contrast to the original I3D design, our prediction is performed for the last frame each sequence continuously adding the most recent frame and removing the least recent frame.  

\subsection{Visual Servoing}
% The visual servoing module does not rely on temporal information.
As illustrated in \figref{fig:approch_block_diagram}, this module uses a lightweight fully-convolutional network to generate control action candidates over a grid of image cells based on an RGB image, corresponding joint angles, and a coordinate tensor. 
The coordinate tensor carries the normalised image UV coordinates of each grid cell, analogous to the approach proposed in ``CoordConv"~\cite{wang2019solo}.
The network architecture is modified based on~\cite{zhuang2020lyrn, zhuang2021multi}. 
This network regresses the value of a control Lyapunov function along with the associated control inputs over a grid of rectangular super-pixels. 
% The non-optimal suppression on the control candidates is achieved by 
Since executing the defined control associated with a given super-pixel decreases the associated Lyapunov value, then choosing the minimal regressed Lyapunov value across the segmentation map performs non-optimal suppression on the control candidates and ensures reliable grasping. 
During grasping execution, the end-effector moves straight down for 3cm before executing a grasp action when the regressed minimum Lyapunov value is lower than a threshold. 

\textbf{Data:}
% The training dataset requires the ground-truth object pose to compute the Lyapunov function and corresponding control.
% Instead of collecting real-world data, we opt to generate pre-labelled synthetic data using simulation.
We replicated the real-world setting in Coppeliasim simulator~\cite{coppeliaSim}. For each sample, the number of simultaneous instances is equally sampled between one and two. The instances are spawned at random poses on the tabletop amongst a collection of random distractors. The end-effector is positioned at a random 6 DoF pose within the workspace of the manipulator.
The generated dataset for IKEA Lack table legs contains approximately 55k samples. 

% \begin{figure}[t]
% 	\centering
% 	\includegraphics[width=0.48\textwidth]{figs/lyrn_pipeline}
% 	\caption{Architecture of the proposed closed-loop reaching algorithm.
% A fully-convolutional network densely predicts a control Lyapunov function (cLf) value $\widehat{\mathcal{V}}$ and control $\widehat{u}$ associated to each foreground image grid cell.
% Non-optimal suppression is achieved by selecting the control associated with the grid cell corresponding to the lowest cLf value.
% The control is updated in real-time as the image and joint angles are updated.
% The reaching trajectory terminates when the regressed Lyapunov value is lower than a threshold.}
% 	\label{fig:lyrn_pipeline}
% \end{figure}

\subsection{Visual handover and system logic}
The process of handing over an assembly part between two humans is often natural, seamless and requires little effort. While humans have multiple sensing modalities that help gather information on this interaction they consciously and unconsciously perform predictions that allow them to adapt to other humans' behaviour. Additionally, humans quickly learn the assembly process which further improves their predictions. Hard coding these elements into a robotic system may be possible in very structured and scripted environments, however, in a flexible, unstructured and dynamic environment, this poses a challenge. GoferBot is able to overcome it by combining the signals from all of its modules and perform a robot-to-human handover.
% To fully understand the handover module we first provide details about the system logic, encapsulated in a finite state machine. 
Note that algorithmic task modelling is outside the scope of this study. 

% \noindent
\textbf{System logic: } The robot starts the assembly in the \textit{home} state. Then the visual servoing module provides a control vector and a termination signal which transitions the system into either the \textit{reach and grasp} state or the \textit{finished} state (when the assembly finishes). During the \textit{reach and grasp} state, the robotic arm moves according to the inferred control vectors, grasps and picks up the assembly piece. Once picked up, the system transitions into the \textit{pass} state where the robotic arm moves into the handover location and transitions into the \textit{idle} state where the robot is essentially waiting for the human to complete their current sub-assembly action. Throughout these states, the action recognition module continually passes the current human action, providing the system with the indication of the assembly progress. Once two consecutive human grasp actions were observed, the system transitions to the \textit{handover} state where the gripper releases the part and the system transitions back to \textit{home} state.  This repeats until assembly commences. 

% \noindent
\textbf{Visual handover: } 
The visual handover is a critical point where all system modules interact. GoferBot's visual handover module receives the action prediction from the action recognition module. The \textit{human reach} atomic action is indicative of an imminent \textit{human grasp} atomic action. 
Since the action prediction is somewhat noisy, the handover module waits until it gathers two consecutive grasp predictions which then trigger the grasping module to release the item.

\section{GoferBot Evaluation}
As an HRC system, we evaluate GoferBot's performance in terms of both robot-centric and human-centric measures.
For the robot-centric measure, we focus on analysing the failure mode of each sub-module and the full system as a whole (Sec.\ref{subsec: robot_centric}).
For the human-centric evaluation, we investigate the human efficiency and experience in Sec.\ref{subsec:hrc_eval}. 

\subsection{Experiment Design}
We conduct two sets of experiments, one focuses on evaluating the visual handover module, and the other on the full assembly. 
The handover task is specified as a human grasp followed by a robot release, and the full assembly is specified by four single-leg assembly sequences: the robot reaches and grasps a leg, the robot moves to the delivery point, the robot waits for the human to complete assembly, human reaches and grasps the leg, the robot recognises a human grasp and releases the leg.

In both sets of experiments, participants were asked to perform the task twice, once with vision-based handover and once with voice command handover as the baseline.
The starting handover mode was chosen randomly. In the voice-command system, we mimic an ideal voice recognition system by having the supervisor press a button to release the assembly part as soon as the command was issued by the participant. The human operation has a zero failure rate and unperceivable response latency. 
We chose this ``ideal voice command" system as a baseline because it is semantic and perception-based.
A similar baseline may be achieved by using a torque sensor, however, this would be an unfair comparison since the sensor contains non-semantic information.

For the handover experiment, each participant performs five handover repetitions in each mode and in the full assembly the participant aims at completing one full assembly.
The system resets in the full assembly experiment if the visual servoing module fails to grasp a part or the action recognition module persistently fails to trigger a release.

Both experiments were conducted with the same group of 10 participants. The participants were 20-30 years old with a background in technology (engineering or computer science) with a 5:1 male:female ratio. 

\subsection{Robot-centric Evaluation}
\label{subsec: robot_centric}
\textbf{Failure mode evaluation: }
A full assembly requires four consecutive leg assembly cycles. Each cycle has two failure modes --- one associated with the visual servoing module failing to grasp and the other associated with the action recognition module failing to detect that the human has grasped the object.
The failure of the visual handover module does not necessarily lead to the failure of a grasp-handover cycle, since the false negative (refusal to release) might be resolved by the human repeating the reach and grasp action.

In \tabref{tab:success_results}, we report the number of failure modes observed during each individual leg assembly cycle as part of the visual full assembly experiment. In summary, there are 50 leg assembly cycles collected from 13 full assembly attempts. A second handover attempt is performed by the participant in 4 of the 50 leg assembly cycles, hence 54 handover attempts. 
The visual servoing module and the visual handover module achieves 96\% and 85.1\% success rate, respectively. 
In the isolated handover experiment, the visual handover module failed 8 times out of 55 trials (85.4\% success rate), this is consistent with the performance in the full-assembly task.
The success rate of completing a reach-grasp-handover cycle is 90\% when allowing multiple human handover attempts. 

\textbf{Action recognition evaluation: }
% \subsection{Action recognition evaluation}
To independently evaluate the performance of the action recognition system we perform a 14-fold cross-validation train-test splits on the collected data, each split contains 4 training videos and 2 test videos. For each split we train a model and measure its per-frame accuracy and mean average precision (mAP) on the test videos and report the results of  $0.67\pm0.05$ and $0.37\pm0.06$ respectively. It shows that I3D performance is still far from ideal, however, the performance is higher than expected when comparing to the IKEA ASM baseline performance (accuracy of 57.58 and mAP of 28.77). Per-class performance in \tabref{tab:results:per_action_performance} shows high scores for visually distinct actions that appeared in the large IKEA ASM dataset (\eg flip tabletop, spin leg) and lower scores for shorter, and less visually distinct actions, some of which only appeared in our fine-tuning dataset (\eg reach and human grasp).

While the performance is not high on a frame-by-frame basis, it is adequate when integrated into an end-to-end system because of the continual stream of frames. Furthermore, due to the data imbalance, the model is biased towards predicting a \textit{no assembly action} label. In the system implementation, we mitigate this by scaling down the associated confidence. 
During our full assembly experiments, we observed that the action recognition system exhibits a performance decrease due to human behaviour that was not observed in the training set, \eg human reaching to grasp after adjusting glasses, created an out-of-distribution motion.
The impact of the occasional unresponsiveness of the visual handover module over the human-robot collaboration experience is studied in the human-centric evaluation in the next section.

% \begin{table}[]
%     \centering
%     \begin{tabular}{c c c}
%     \toprule
%           &  \textbf{Accuracy} & \textbf{mAP}\\
%          \hline
%             Action recognition  &$0.67\pm0.05$&$0.37\pm0.06$\\
%          \bottomrule
%     \end{tabular}
%     \caption{Action recognition performance - 14-fold cross validation per-frame accuracy and mean average precision.}
%     \label{tab:results:cross_val_perception}
% \end{table}

\subsection{Human-centric Evaluation}
\label{subsec:hrc_eval}

\begin{table}
    \vspace{2mm}
    \centering
    \begin{tabular}{c c c c}
	\toprule
          \textbf{Action}&\textbf{precision}&\textbf{recall}&\textbf{f1-score}\\
			\midrule
            no assembly action &0.72&0.80&0.75\\
            reach&0.40&0.13&0.14\\
            flip table top&0.79&0.80&0.79\\
            flip table&0.64&0.52&0.53\\
            spin leg&0.79&0.61&0.66\\
            align leg&0.48&0.37&0.40\\
            rotate table&0.66&0.56&0.56\\
            human grasp&0.27&0.17&0.13\\
            \midrule
            Average&0.59&0.49&0.50\\
         \bottomrule
    \end{tabular}
    \caption{Per action recognition performance.}
    \label{tab:results:per_action_performance}
    % \vspace{-4mm}
\end{table}

\begin{table}
   % \vspace{2mm}
	\begin{center}
	    \begin{tabular}{c c c c}
% 		\begin{tabular}{M{0.15\linewidth} M{0.15\linewidth} M{0.15\linewidth} M{0.15\linewidth} M{0.15\linewidth}}
		
			\toprule
% 			& \textbf{\makecell{Robot-Grasp}}& \textbf{Handover} & \textbf{\makecell{R-Grasp+\\Handover}} &\textbf{\makecell{Full-\\Assembly}}\\
    & \textbf{Robot-Grasp} & \textbf{Handover} & \textbf{Cycle}\\
% % 			\midrule
% 			 &  & \smaller F/P & \smaller F/N & \smaller  \\
			\midrule
			\textbf{Fails/total}
			& 2/50 & 8/54 & 5/50 \\
% 			\textbf{Attempts}& 50 & 54 & 50 & 13\\
			\midrule
			\textbf{\makecell{Success Rate}}& 96\%& 85.1\%& 90.0\% \\
			\bottomrule
		\end{tabular}
	\end{center}
	\vspace{-2mm}
	\caption{Success rate from the full-assembly experiments.} 
% 	Multiple human grasp attempts are allowed to initiate the handover, if false negative (refusal of release) present in the visual handover module.
	\label{tab:success_results}
\end{table}

\textbf{Hypothesis 1 (H1) --- Handover efficiency:} \textit{Vision-based handover is more efficient than command-based handover measured by a reduction in handover completion time.}

\begin{figure}[t]
    \vspace{2mm}
	\centering
	\setlength\tabcolsep{1pt}
	    \begin{tabular}{c c | c c}
	                \multicolumn{4}{c}{
            \includegraphics[width=0.5\linewidth]{figs/boxplot_legend.png}}\\
            \small{\rotatebox[origin=c]{90}{Time [s]}} &
            \begin{subfigure}{.18\linewidth}
                \centering
                \includegraphics[width=0.9\linewidth]{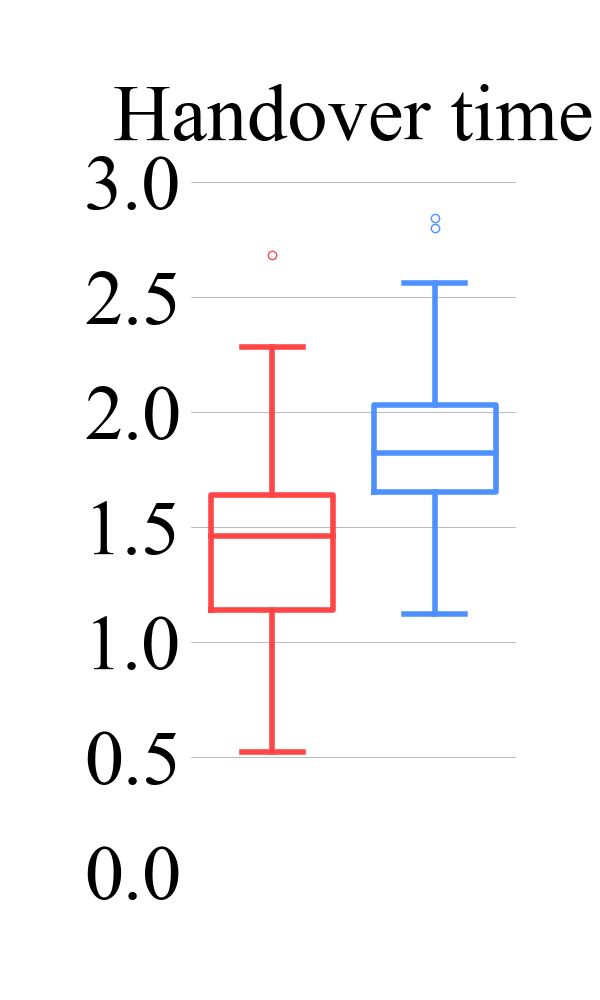}
                \caption{}
                \label{subfig:handover_total_time}
            \end{subfigure} 
            \begin{subfigure}{.18\linewidth}
                \centering
                \includegraphics[width=0.9\linewidth]{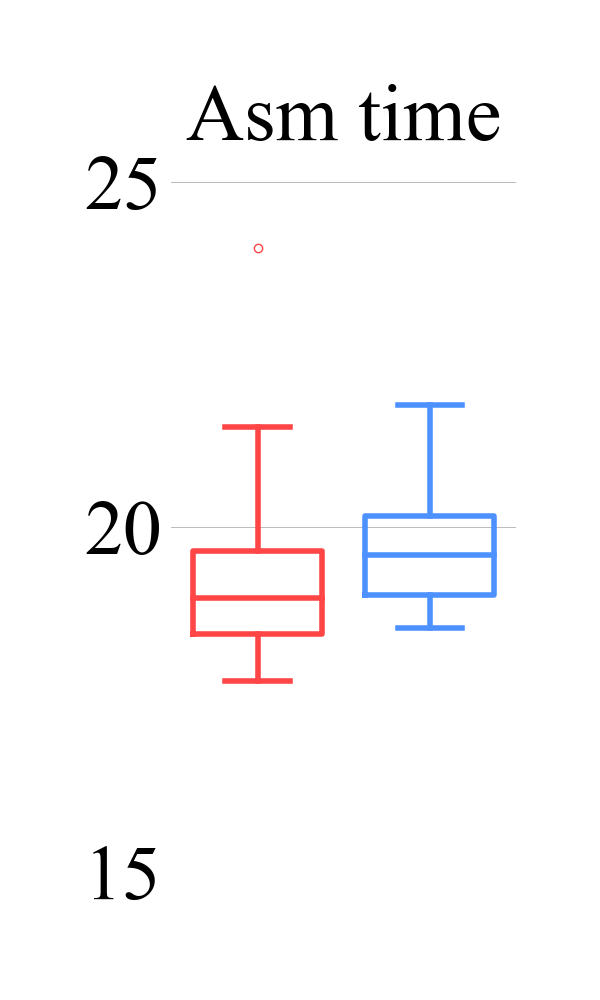}
                \caption{}
            \end{subfigure}
            &\small{\rotatebox[origin=c]{90}{ratio [\%]}} &
                        \begin{subfigure}{.18\linewidth}
                \centering
                \includegraphics[width=0.9\linewidth]{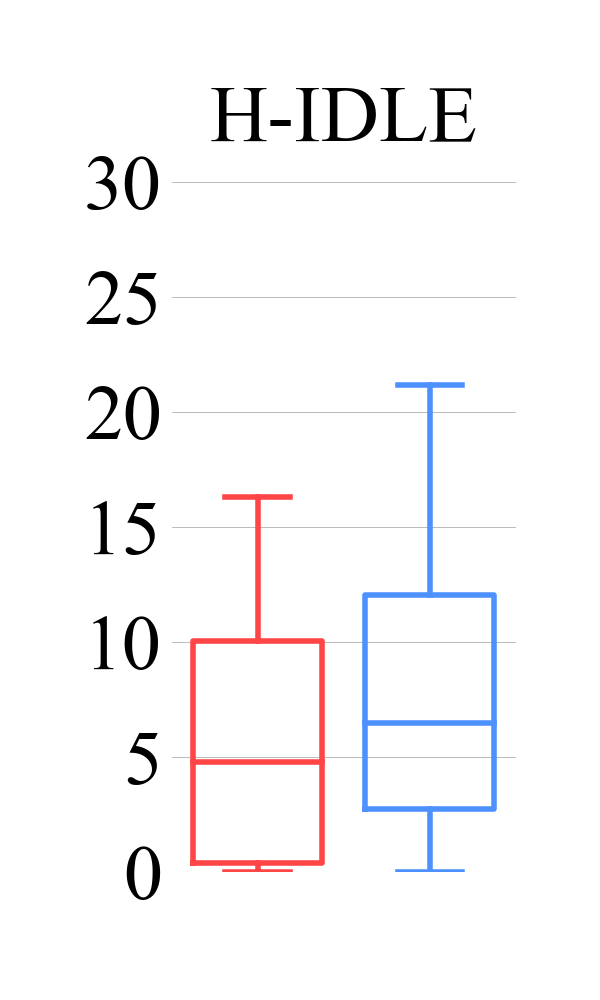}
                \caption{}
            \end{subfigure}
                                    \begin{subfigure}{.18\linewidth}
                \centering
                \includegraphics[width=0.9\linewidth]{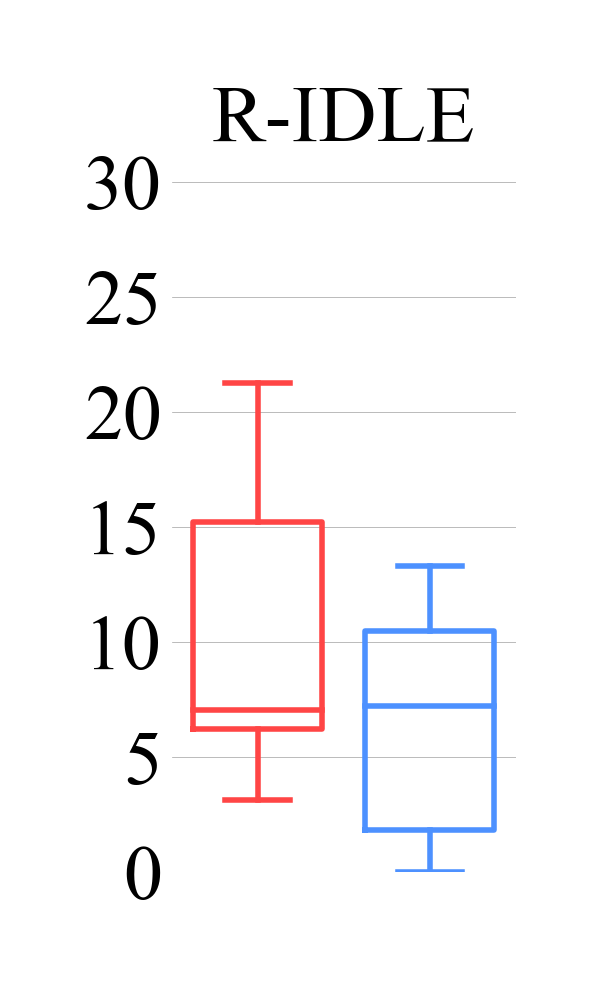}
                \caption{}
            \end{subfigure}
                                    \begin{subfigure}{.18\linewidth}
                \centering
                \includegraphics[width=0.9\linewidth]{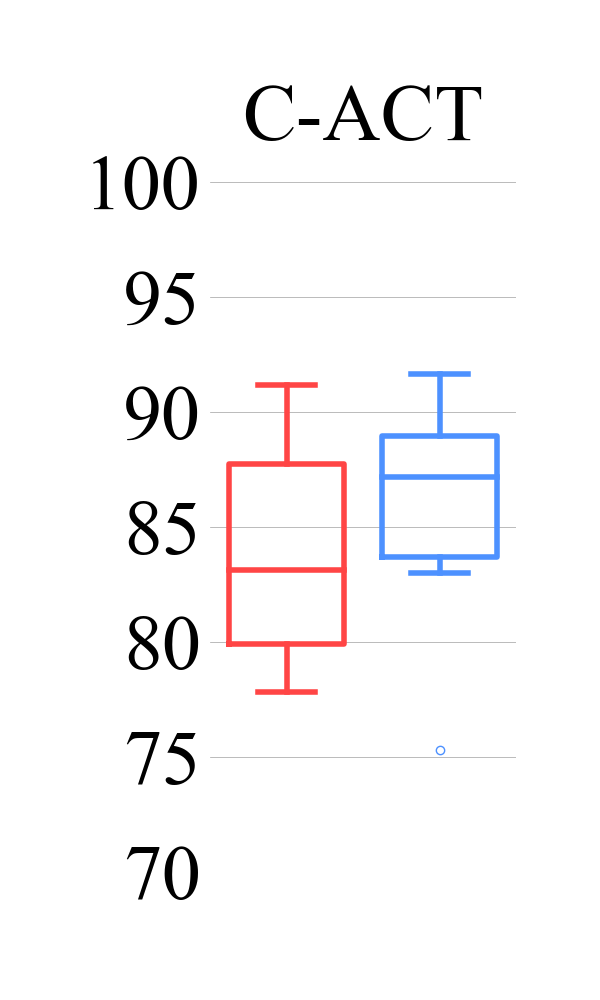}
                \caption{}
            \end{subfigure}\\

    	    \end{tabular}
    
	\caption{Objective metrics results for the handover and full system performance. (a) Handover total time (b) Single leg assembly cycle time, (c) Human idle time ratio, (d) Robot idle time ratio, (e) Concurrent action time ratio.}
	\label{fig:resutls:handover_time}
	\vspace{-2mm}
\end{figure}

\noindent
In the handover experiment, we measure the total handover time. This is measured from the moment a human initiates a grasp action (hand closure) until the start of object release. The moment a human initiates a grasp for the voice command baseline is the moment the command is issued while in the vision based system it is at the end of the reaching action (when the human touches the part). 
We report the single-leg assembly cycle time measurements, computed as the average of all successful cycles per participant. The time measurements are presented in \figref{fig:resutls:handover_time}(a). Since the collected time data was not normally distributed, we conduct the Wilcoxon signed-rank test~\cite{wilcoxon1992individual} and found that there was a statistically significant difference between the measured time distributions \ie the vision-based handover is faster than the command based by approximately 0.5 seconds on average ($\sim27\%$ faster). Therefore, we accept the H1 hypothesis. 

\textbf{Hypothesis 2 (H2) --- Assembly efficiency:} \textit{Vision-based assembly is as efficient as command-based assembly, measured by robot/human idle and concurrent action time.}

\noindent
In the full assembly experiment, we show that, despite having multiple modules performing simultaneous perception tasks, there is no increase in assembly time compared to the ideal voice command baseline. While the handover was shown to be faster, its duration is relatively short compared to all other actions within the assembly cycle and would have a small effect on the overall assembly task. In \figref{fig:resutls:handover_time}(b), we report the human idle time ratio (H-IDLE), robot idle time ratio (R-IDLE) and concurrent action time ratio (C-ACT) within the assembly cycle.  We found that there was no statistically significant difference between the vision and voice command system modes and therefore accept the H2 hypothesis. Note that this is quite remarkable since the voice command baseline is partially operated by a human to mimic human-like response times.

\textbf{Hypothesis 3 (H3)--- Handover/assembly experience:} \textit{Vision-based handover improves the overall experience of the handover compared to command-based handover, measured by subjective measures.}

\noindent
For the subjective measures, we adopt the Likert items and scales as in Han and Yanco~\cite{han2019effects}, given here in \tabref{tab:likert_questions}. We reverse the responses about discomfort to comfort and report a Cronbach's alpha at an acceptable level of internal consistency reliability ($\alpha=0.76$). We performed a pairwise Wilcoxon signed-rank test \cite{wilcoxon1992individual} and report that for both the full assembly experiment and the handover experiment.

\begin{figure}[t]
	\centering
	\setlength\tabcolsep{1pt}
	    \begin{tabular}{c c c}
	                &\multicolumn{2}{c}{
            \includegraphics[width=0.5\linewidth]{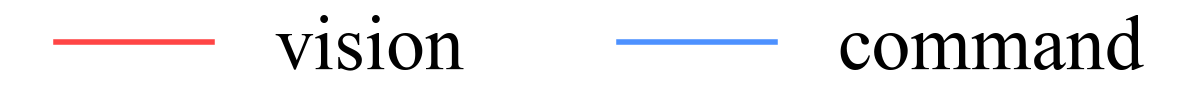}} \\
            (a) &\rotatebox[origin=c]{90}{Likert ratings} &
            \begin{subfigure}{.18\linewidth}
                \centering
                \includegraphics[width=0.9\linewidth]{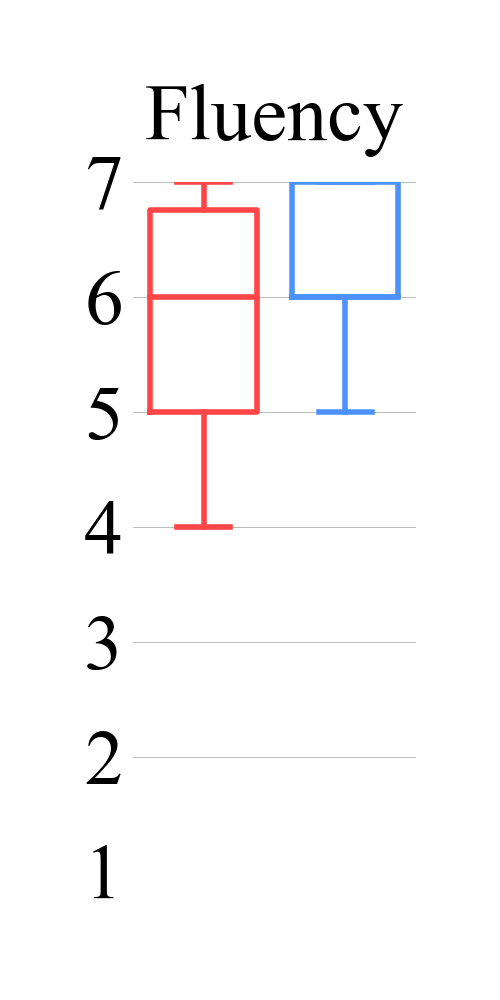}
            \end{subfigure}
                \begin{subfigure}{.17\linewidth}
                \centering
                \includegraphics[width=0.9\linewidth]{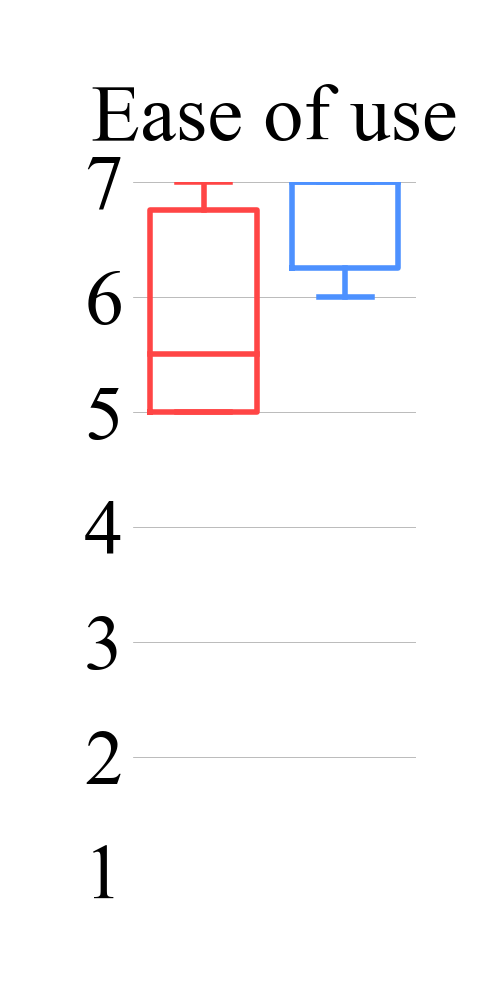}
            \end{subfigure}
                \begin{subfigure}{.17\linewidth}
                \centering
                \includegraphics[width=0.9\linewidth]{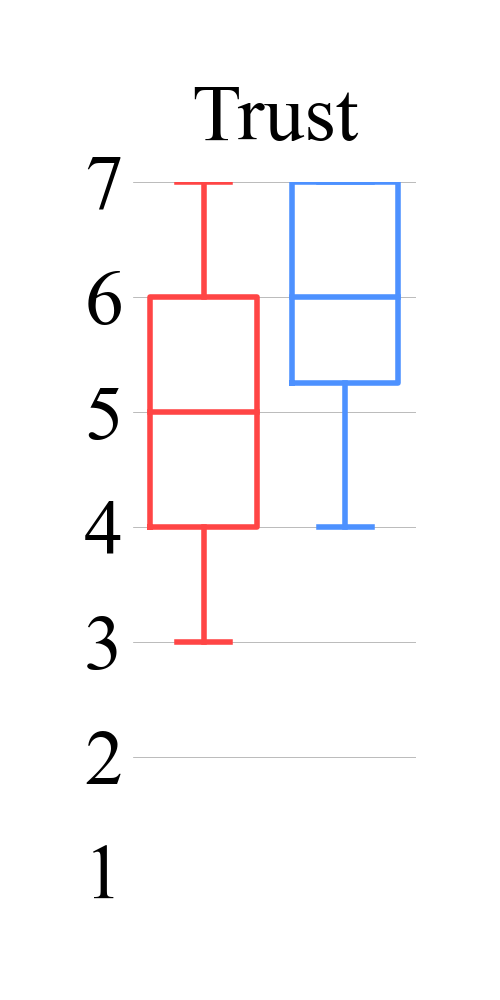}
            \end{subfigure}
            \begin{subfigure}{.17\linewidth}
                \centering
                \includegraphics[width=0.9\linewidth]{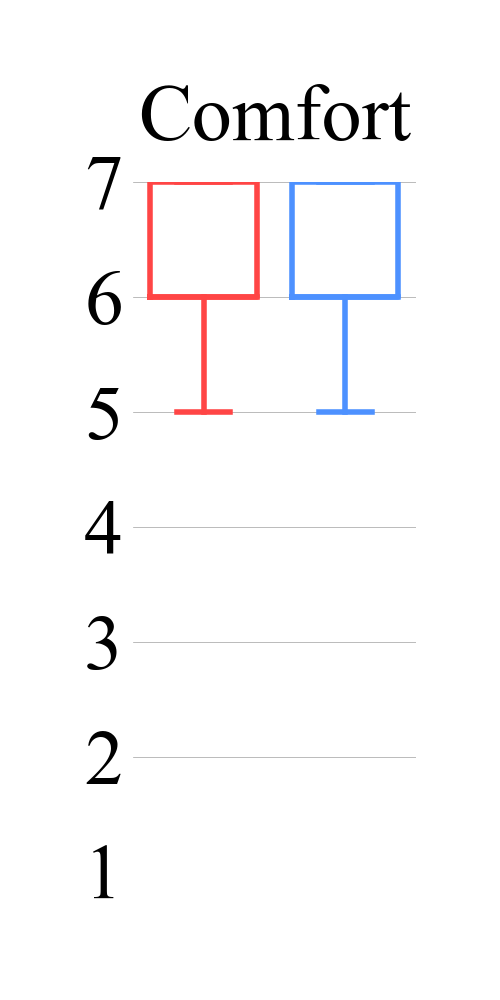}
            \end{subfigure}
            \begin{subfigure}{.17\linewidth}
                \centering
                \includegraphics[width=0.9\linewidth]{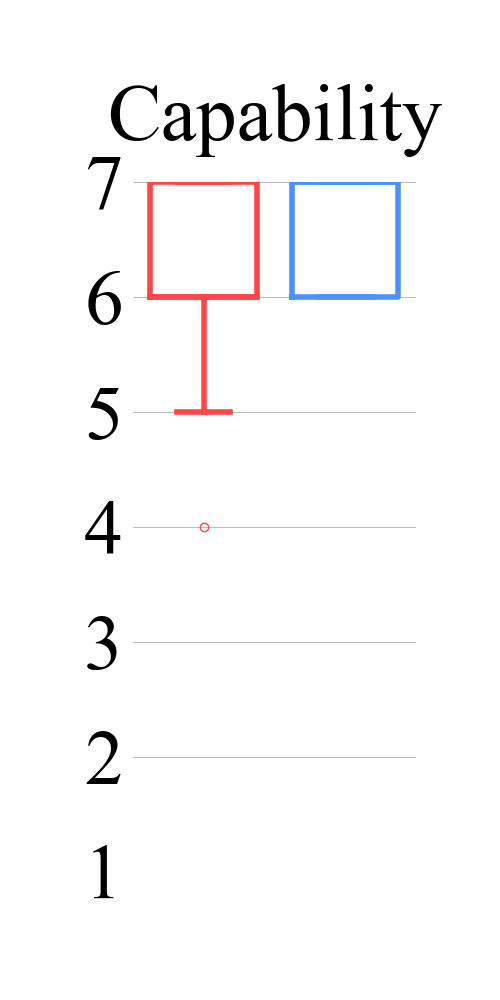}
            \end{subfigure} \\
                        (b) &\rotatebox[origin=c]{90}{Likert ratings} &
            \begin{subfigure}{.18\linewidth}
                \centering
                \includegraphics[width=0.9\linewidth]{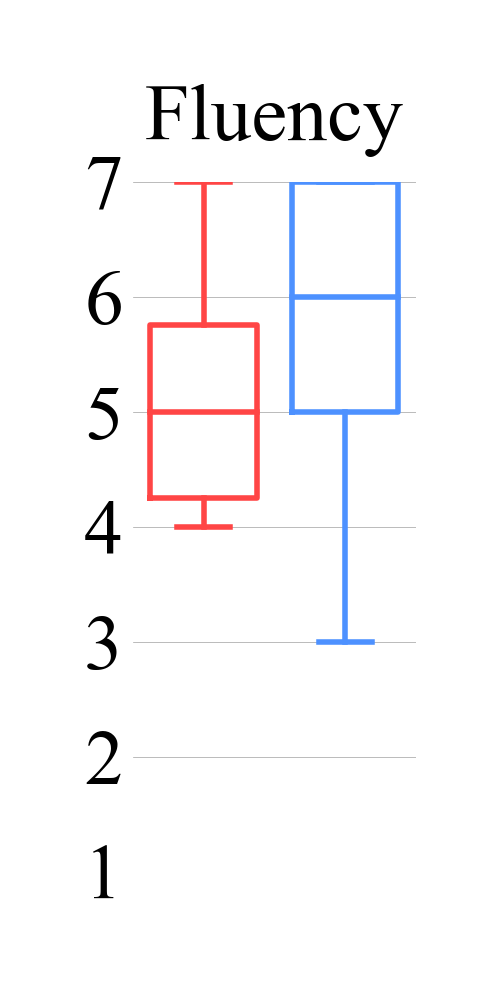}
            \end{subfigure}
                \begin{subfigure}{.17\linewidth}
                \centering
                \includegraphics[width=0.9\linewidth]{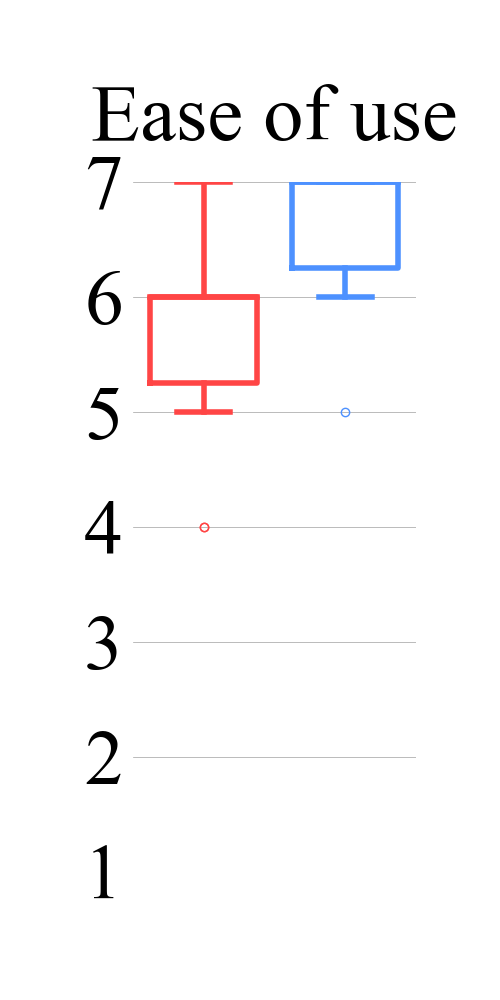}
            \end{subfigure}
                \begin{subfigure}{.17\linewidth}
                \centering
                \includegraphics[width=0.9\linewidth]{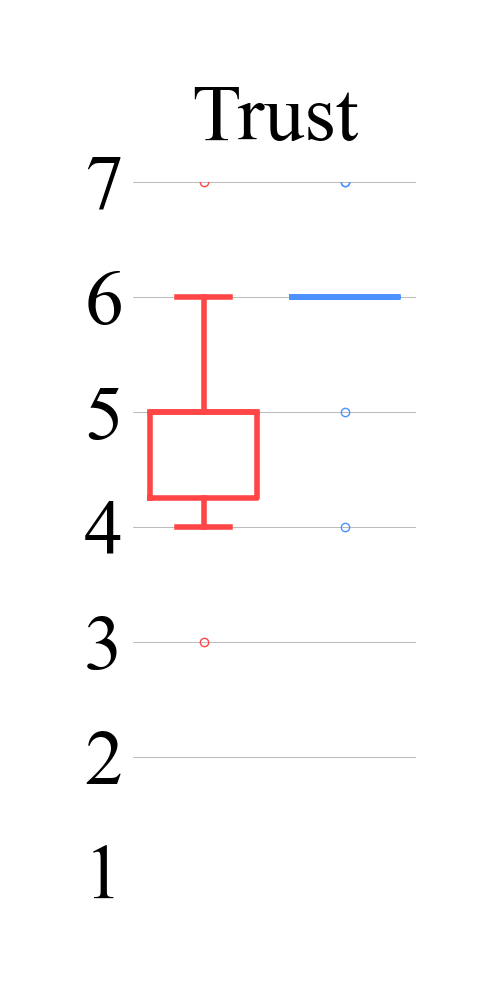}
            \end{subfigure}
            \begin{subfigure}{.17\linewidth}
                \centering
                \includegraphics[width=0.9\linewidth]{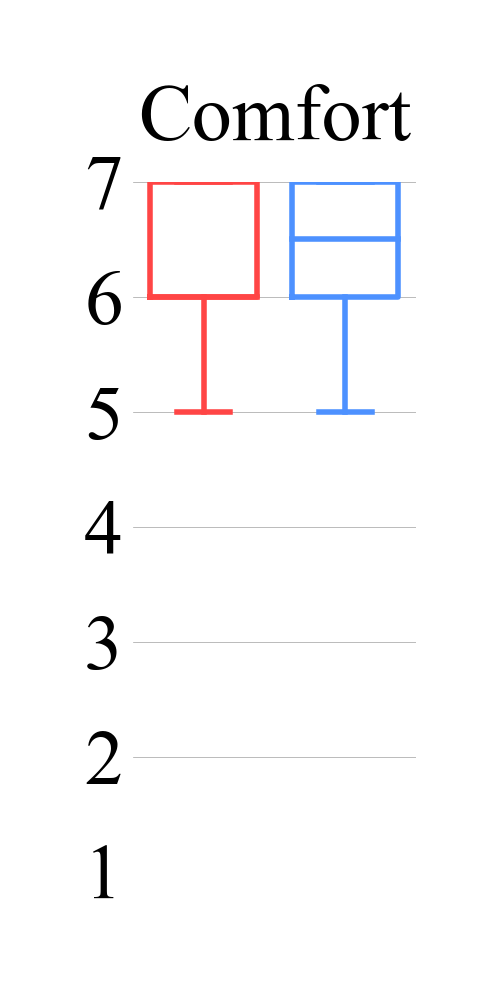}
            \end{subfigure}
            \begin{subfigure}{.17\linewidth}
                \centering
                \includegraphics[width=0.9\linewidth]{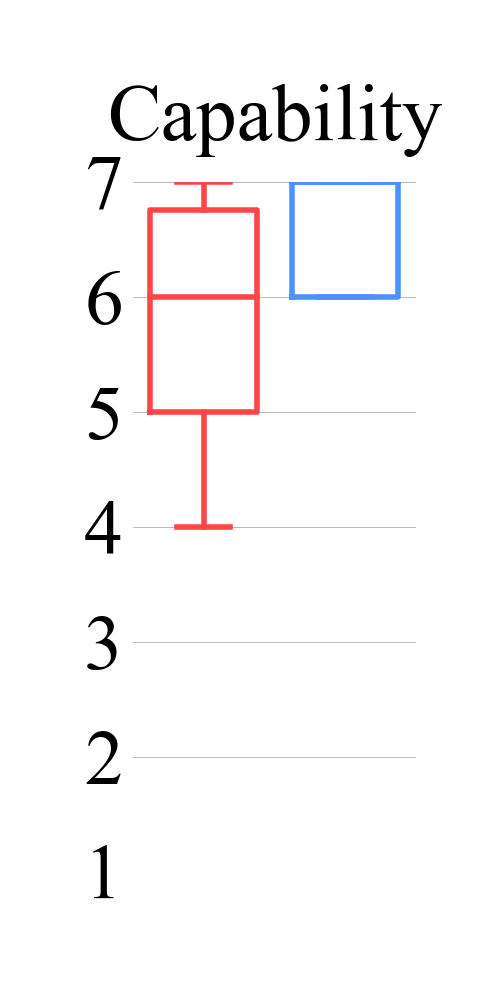}
            \end{subfigure} \\
            &\multicolumn{2}{c}{Likert Items} \\

    	    \end{tabular}
    
	\caption{Questionnaire responses summary as box plots for (a) full assembly experiment, and (b) handover experiment.}
	\label{fig:results:likeart_scores}
\vspace{-4mm}
\end{figure}

As shown in \figref{fig:results:likeart_scores}, there is a statistically significant difference in ease of use and trust. The participants find the command-based system method easier to use and more trustworthy compared with a vision-based approach.
The lower score in the trust and ease-of-use mainly originates from experiments that exhibit unresponsiveness of the handover sub-module. Note that the command-based handover is ideal and does not contain any misdetections or unresponsiveness.

No statistical significance is observed in fluency, comfort and capability. This indicates the participants find both command-based and vision-based approaches' capability, fluency and comfort to be equivalent. 
This result contradicts the objective measures of handover efficiency.
It can be attributed to the humans' perception of control --- the vision-based system provides no feedback for the system internal execution state while in the command-based system, the human determines the system state with their command.  
The overall Likert ratings are reported in a box plot in \figref{fig:results:likeart_scores}.
Therefore, we reject the H3 hypothesis for both the handover task and the full assembly task. 
 
\begin{table}
    \vspace{2mm}
	\begin{center}
		\begin{tabular}{l}
			\toprule
            \textbf{Fluency} (Cronbach's $\alpha = 0.79$ if deleted)\\
            \textit{The robot handed over like a human.}\\
            \hline
            \textbf{Ease-of-use} (Cronbach's $\alpha = 0.62$ if deleted)\\
            \textit{It was easy to take objects from the robot.}\\
            \hline
            \textbf{Trust} (Cronbach's $\alpha = 0.63$ if deleted)\\
            \textit{I trust the robot to do the right thing at the right time.}\\
            \hline
            \textbf{Comfort} (Cronbach's $\alpha = 0.79$ if deleted)\\
            \textit{I feel uncomfortable with the robot. (reversed rating)}\\
            \hline
            \textbf{Capability} (Cronbach's $\alpha = 0.7$ if deleted)\\
            \textit{The robot was capable of handing over the object.}\\
            \hline
            \small{Note: Likert items are coded as 1 (strongly disagree), }\\
            \small{ 2 (disagree),3 (moderately disagree), 4 (neutral),  }\\
            \small{5 (moderately agree), 6 (agree), 7 (strongly agree). }\\
			\bottomrule
		\end{tabular}
	\end{center}
	\caption{Likert scale questionnaire ($\alpha = 0.76)$}
	\label{tab:likert_questions}
	\vspace{-3mm}
\end{table}

% \textbf{Human-perceived time and objective fluency.}

% 1. frame rate used to train action-recognition model.
% 2. failure rate. 
% 3. Compared with mid to large-scale image datasets that contain more than a few million images, manufacturing related dataset are comparatively small that can not cover the distribution of real-world scenario, hence the gap between the accuracy achieved by evaluation datasets and real-world experiments.

% In this paper, we  demonstrated that the proposed GoferBot works well in the context of an IKEA table assembly.
% The system has several limitations:
% First, since GoferBot is a purely vision based system, significant changes to lighting, and human appearance may affect performance. 
% This limitation can be resolved by increasing quantity and diversity of the training set. 
% Second, changing assembly tasks requires an additional task modeling component. While the action recognition and visual servoing modules are more than capable of facilitating a large number of different actions and objects, the finite state machine needs to be adapted for different tasks either manually or algorithmically (which is outside the scope of this work). 
% Finally, our evaluation focused on quantifying the system's efficiency and user experience, however there are other aspects of human-robot collaboration that can be evaluated and are outside the scope of this paper, including proactive handover strategies, efficiency of robot trajectories, and task modeling, 

\section{Limitations}

% This section discusses three key aspects around designing integrating data-driven vision-based approaches into a full human-robot collaboration system.
% \begin{itemize}
%     \item Process frame rate and data distribution;
%     \item The non-indicative dataset-based evaluation for real-world performance;
%     \item Exponential decay effect for repetitive tasks.
%     \item Human-perceived time and objective fluency.
% \end{itemize}

% \textbf{Real-world against dataset-based evaluation:}
% Large-scale computer vision dataset covers a larger range of data distribution, when testing in the same domain, the performance difference degradation is small. Can be collected from the internet.
% However, for human-robot interaction-based dataset, which has considerably small data size collected from limited participants, covering less diverse scenario, is subject to reduced generalisability.  
% From the experiments we performed, although the evaluation results based on dataset statistics is considerablly low, the real world performance can be sufficiently mitigated by multi-frame connects at the cost of reaction time.

\textbf{Exponential performance decay for repetitive tasks:}
Sub-task repetition is fundamental to assembly and manufacturing.
While Goferbot achieves a considerably high success rate ($90\%$) based on 50 real-world ``robot-grasp to human-assemble" cycles, the system performance faces the challenge of exponential decay due to task repetitions.
In our test scenario, a full table assembly requires four consecutive successful leg assembly cycles, assuming each assembly cycle is an independent event, the probability of having a successful full assembly is $0.9^4=0.656$, which broadly agrees with the observed success rate of 61.5\%.
To achieve $90\%$ full-assembly success, Goferbot needs to reach approximately $97.4\%$ individual task success rate.
Although this goal may not be immediately attainable for visual learning-based real-world human-robot assembly tasks, we think it is critical to bring the task repetition performance degradation into attention --- isolated experiments of individual modules are non-indicative for system performance on repetitive tasks and future work should focus on system-level evaluation. 

\textbf{Robustness to input frame rate variation}:
It is crucial to consider the robustness of frame-rate variation while developing a real-world visual learning-based system. Most action recognition approaches rely on having a constant-size image buffer. Therefore, the frame rate dictates the network's learnt prior about the speed of an action. To mitigate this, Goferbot's action recognition network was trained and tested on similar frame rates. Future work should focus on frame-rate independent action predictions.     

% The history available to Goferbot's action recognition module is dependent on the sampling rate of the original input video stream and the size of the image buffer. 
% For example, when the input 30Hz video is sampled at 30Hz and 10Hz, the selected 16-frame buffer can hold the most recent 0.53 and 1.6 seconds, respectively. 
% Keeping a relatively long history is especially advantageous for recognising extremely short atomic actions such as human grasp (hand closure) as the additional history provides insight into the transitions from its preceding action. 
% However, changing the input frame rate affects the network's learnt prior about the speed of an action.
% We observed a considerable performance reduction when the network is deployed at one-third the input frame rate used for training.
% The final action recognition network is fine-tuned at the buffering frequency (10Hz).
% The visual grasping algorithm used in this work is single-frame based that is naturally robust to frame-rate variation (within reasonable closed-loop frequencies).

\section{Conclusion}
We present a novel visual learning-based, grasp-to-handover human-robot collaborative system in a dynamic and unstructured assembly scenario. 
% We demonstrate that the feasibility and importance of using monocular vision as the only sensing modality to create non-intrusive and intuitive collaboration behaviour.
The proposed system develops natural implicit semantic understanding and communication without intervening with the environment e.g. Fiducial markers or tracking devices.
GoferBot shows the feasibility and importance of a vision-based approach for smart manufacturing.
Based on both human and robot-centric evaluations, we show that the system is able to achieve $90\%$ accuracy on leg assembly cycles and objectively faster human-robot interaction than the ideal voice-command baseline. This work opens new interesting directions for future work. The action recognition module, for example, currently tries to mimic a human's perception of action and it may benefit from an integrated forecasting ability that humans exhibit naturally as well as a subtle feedback module that humans provide through gaze and body language.

% This work opens new interesting directions for future work, including improving each of the individual modules' performance --- action recognition, visual servoing for grasping and visual handover. 
% Furthermore, while the action recognition module currently tries to mimic a human's perception of action it may benefit from an integrated forecasting ability that humans exhibit naturally as well as subtle feedback that humans provide through gaze and body language. 
% While improving the performance of each individual module can be further explored  in future work, our evaluation suggests that isolated experiments of individual modules are non-indicative for system performance on repetitive tasks and future work should focus on system-level evaluation. 
% % Future work should focus on We observed that isolated experiments are non-representative of the full task 
% Future work may also improve the system's performance by introducing redundancies in the form of a force sensor on the gripper that will provide an additional modality for state prediction. Other modalities of interest include depth information that can help reduce lighting effects.  

%\section{Acknowledgements}
\noindent\textbf{Acknowledgements}: This work was supported by The Australian Centre for Robotic Vision and the European Union’s Horizon 2020 research and innovation programme under the Marie Sklodowska-Curie grant agreement No 893465. 

%% Use plainnat to work nicely with natbib.
% \clearpage
\bibliographystyle{IEEEtranS}
\bibliography{ref_full}

\end{document}